\documentclass[10pt,twocolumn,letterpaper]{article}
\usepackage{iccv}
\usepackage{times}
\usepackage{epsfig}
\usepackage{graphicx}
\usepackage{amsmath}
\usepackage{amssymb}
\usepackage{epstopdf}
\usepackage{xcolor}
\usepackage{comment}
\usepackage[accsupp]{axessibility}
\pdfoutput=1

\usepackage[breaklinks=true,bookmarks=false]{hyperref}

\iccvfinalcopy 

\usepackage{times}
\usepackage{epsfig}
\usepackage{graphicx}
\usepackage{float}
\usepackage{wrapfig}
\usepackage{amsmath,amssymb,amsthm}
\usepackage{algorithm,algorithmicx,algpseudocode}
\usepackage{bm,xspace}
\usepackage{comment}
\usepackage{verbatim}
\usepackage{multirow}
\usepackage{balance}
\usepackage{url}
\usepackage{booktabs}
\usepackage{etoolbox,siunitx}
\usepackage{calc}
\usepackage{pifont,hologo}
\usepackage{nicefrac}

\newcommand{\ra}[1]{\renewcommand{\arraystretch}{#1}}

\usepackage[super]{nth}
\usepackage{nicefrac}
\robustify\bfseries

\DeclareMathSymbol{@}{\mathord}{letters}{"3B}

\def\latex/{\LaTeX}
\def\bibtex/{\hologo{BibTeX}}

\def\iccvPaperID{****} 

\ificcvfinal\fi

\begin{document}

\title{Embedding Novel Views in a Single JPEG Image}

\author{
Yue Wu$^*$ \qquad Guotao Meng\thanks{Joint first authors} \qquad Qifeng Chen\\
The Hong Kong University of Science and Technology\\ }

\maketitle
\ificcvfinal\thispagestyle{empty}\fi

\begin{abstract}
   We propose a novel approach for embedding novel views in a single JPEG image while preserving the perceptual fidelity of the modified JPEG image and the restored novel views. We adopt the popular novel view synthesis representation of multiplane images (MPIs). Our model first encodes 32 MPI layers (totally 128 channels) into a 3-channel JPEG image that can be decoded for MPIs to render novel views, with an embedding capacity of 1024 bits per pixel. We conducted experiments on public datasets with different novel view synthesis methods, and the results show that the proposed method can restore high-fidelity novel views from a slightly modified JPEG image. Furthermore, our method is robust to JPEG compression, color adjusting, and cropping. Our source code will be publicly available.
\end{abstract}

\section{Introduction}

\begin{figure}[t!]
\hspace{-3mm}
\includegraphics[width=1.08\linewidth]{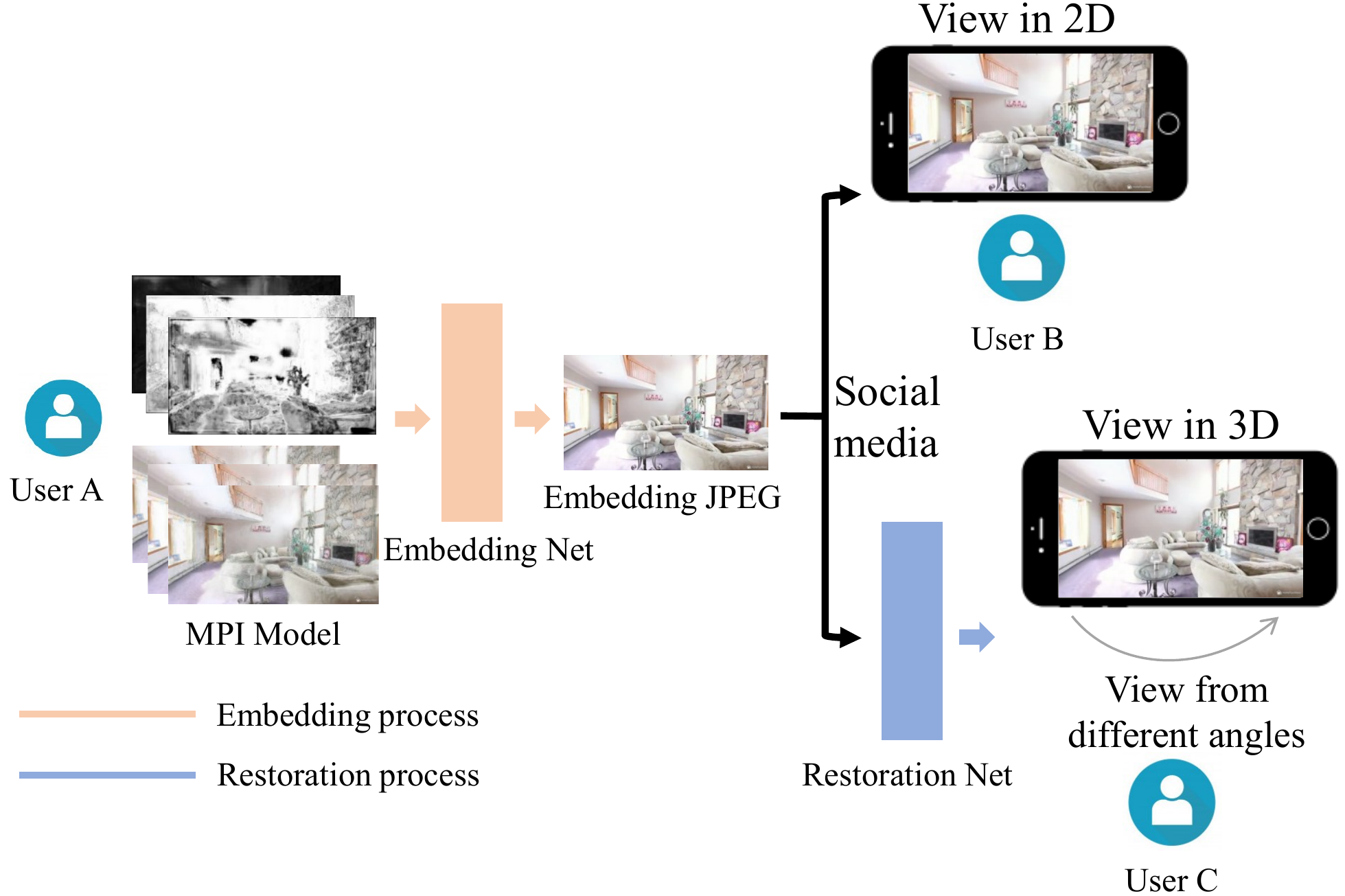}
\caption{An application of our method. Our approach can embed an MPI into a JPEG image, and then share it on social networks. With our restoration network (as a plugin), users can view this image in 3D from different viewpoints. Meanwhile, users without our restoration network can still view this image as an ordinary 2D image.} 
\label{fig:application}
\end{figure}

Novel view synthesis allows users to view a photograph in 3D from different viewpoints, which creates a more immersive experience than a 2D image. While the performance of novel view synthesis models has improved significantly in recent years, sharing the 3D photo rendered by novel view synthesis on social networks remains unsolved yet. As shown in Fig.~\ref{fig:application}, how can User A share a novel view synthesis model with others on social networks? One plausible strategy is to embed the novel synthesis model (MPI model) into a single embedding image (in JPEG format) that looks like an ordinary image. When User B receives the embedding image shared by User A, User B can view it as an ordinary 2D image; on the other hand, User C with a plugin (a restoration network) can see this image in 3D from the restored novel view synthesis model from the embedding image.

To enable such user-friendly applications, we propose a novel approach that embeds an MPI into a single embedding image that can be converted back to MPI. In this work, we mainly focus on MPI because it is a popular representation in several state-of-the-art novel view synthesis frameworks~\cite{flynn2019deepview,habtegebrial2020semantic_mpi,huang2020semantic, PB,single_view_mpi, zhou2018stereo}. With our proposed approach, users can easily share a 3D photograph via a 2D embedding image on social media (e.g., Facebook and Twitter). The embedding image is simply a standard JPEG image,  which is a compressed image format widely adopted in social media or websites. Our proposed method can be implemented as an extension of web browsers or a lightweight plugin of apps. Users without the plugin can see the scene as a JPEG image, while users with the plugin can view this scene from different viewpoints. Our method saves network traffic and storage space benefited from the small file size of the JPEG format with a high compression rate.

\begin{figure*}[t!]
\centering
\setlength\tabcolsep{2pt}
\includegraphics[width=1\linewidth]{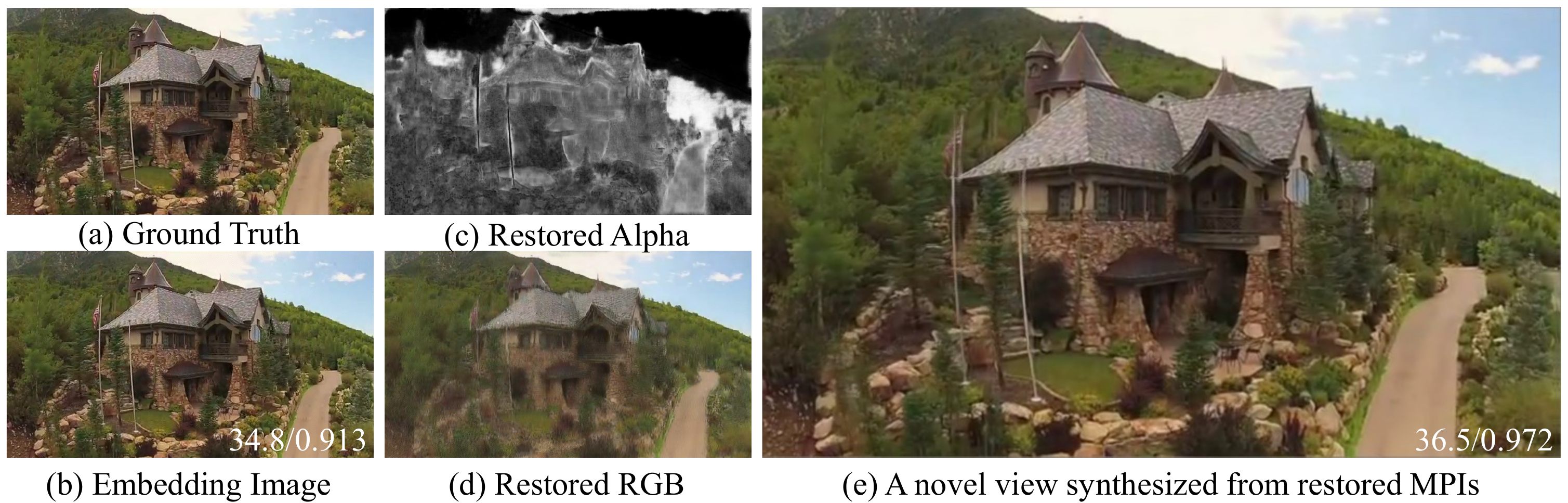}
\caption{(a) is the ground-truth image. (b) is the embedding image that embeds a 32-layer MPI. (c) and (d) are examples of recovered MPI (RGB and Alpha) from (b). (e) is a novel view synthesized from recovered MPI decoded from the embedding image. Our method can achieve high-fidelity novel views while preserving the visual quality of the embedding image. The reported two values in (b) and (e) are PSNR and SSIM.} 
\label{fig:compare_1}
\end{figure*}

Embedding a novel view synthesis model in a single image is a novel and research-worthy task. Compared to steganography, this problem has different objectives and is arguably more challenging. Our objective is not to keep the hidden information undetected. Instead, our objective is to make the embedding image visually pleasing without evident artifacts while the restored novel views are nearly perfect. Existing steganography models commonly hide a short message (a hyperlink~\cite{2019stegastamp}, light field information~\cite{wengrowski2019light}, or a vector~\cite{zhu2018hidden}) and a small number of images~\cite{hu-2020-mononizing, zhu2020video} into a reference image. A direct application of previous techniques in our setting generates low-quality results for embedding images and the restored MPIs.

It is extremely challenging to embed a 32-layer MPI in a JPEG image, because it is equivalent to embed $32 \times 4 \times 8 = 1024$ bits of information into a single pixel. Although the layers in the MPIs are correlated, the MPIs store the content and depth information of multiple views, thus the information amount to be embedded is still high.  Moreover, the JPEG format is ubiquitously used, especially on social media, for its high compression rate. Thus we choose this lossy compression format as the format of our embedding images. 

In our framework, we design a specific neural network architecture, which contains an encoder and a lightweight decoder, based on the property of MPIs. Moreover, we introduce a novel frequency-domain loss and adversarial loss to suppress weird high-frequency artifacts that previous approaches usually have. Besides, because users often retouch the images shared on social media, we apply a set of image perturbations to make our framework robust to real-world scenarios. 

We conduct experiments to evaluate the performance of our proposed model compared to several baselines. Our model significantly outperforms other baselines in novel view synthesis quality and perceptual performance. While the rendered frames in novel views can reach a high fidelity of 36.68 PSNR,  the embedding image can still be preserved similar to the reference image. A visual example is shown in Figure~\ref{fig:compare_1}.

Our contributions can be summarized as follows:
\begin{itemize}
    \item We propose the first dedicated model that can embed MPIs in JPEG images.
    \item We introduce adversarial training and frequency domain loss to suppress high-frequency artifacts in the embedding image.
    \item Our system can embed MPIs in JPEG images in nearly-imperceptible form and restore high-fidelity novel views synthesis. Moreover, our system is robust to a range of image manipulations such as JPEG compression, color adjusting, and cropping.
\end{itemize}

\begin{figure*}[t!]
\centering
\includegraphics[width=1\linewidth]{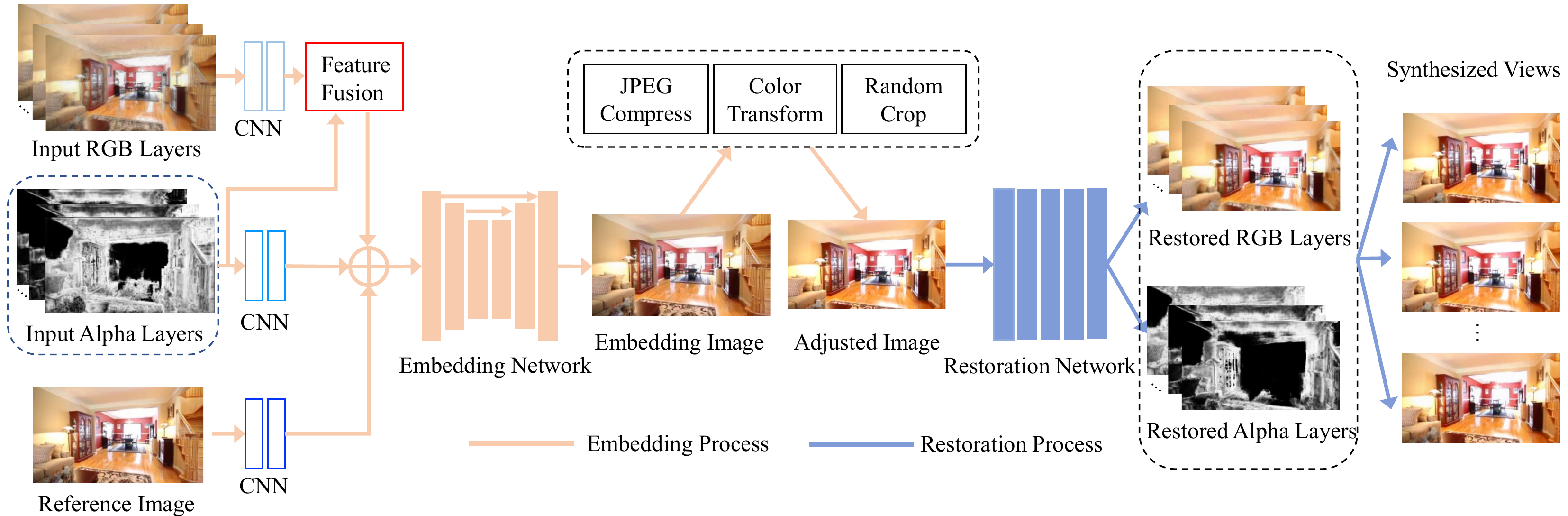}
\vspace{1mm}
\caption{The framework of the proposed method. First, the features of the RGB layers, the alpha layers, and the reference image are fused to feed into the embedding network to generate the embedding image. Subsequently, the embedding image is compressed and altered. Later the restoration network decodes the embedding image into MPIs. Finally, the novel views can be rendered from different viewpoints with the restored MPIs. The whole framework is trained end-to-end.}
\label{fig:overview}
\end{figure*}

\section{Related Work}
\paragraph{Novel view synthesis.}
The research in novel view synthesis is prospering due to the applications of deep neural networks. Novel view synthesis can be formulated as a learning problem. For each scene, some views are used as input, while others are used as target views. The objective of the network is to predict target views based on given views. In these frameworks, the most generally used representation is the multiplane images (MPIs).
Zhou \etal~\cite{zhou2018stereo} firstly propose the MPI scene representation. They propose a deep learning pipeline to train an MPI prediction network using two images as the input. The novel view rendering is accomplished by reprojecting the MPI, consisting of color layers and alpha layers. There are many works following this scheme~\cite{mildenhall2019llff, PB, single_view_mpi,flynn2019deepview,huang2020semantic,habtegebrial2020semantic_mpi, wang2019web, duvall2019compositing}.  LLFF~\cite{mildenhall2019llff} proposes to expand each sampled view into a local light field via MPI representation. Pratul \etal~\cite{PB} focus on generating high-quality view extrapolations with plausible disocclusions using the MPI representation.

Since MPIs are the most commonly used representations for novel view synthesis, we propose embedding the MPI representation into a single RGB image and then using a restoration network to reveal hidden information.

\paragraph{Steganography.}
Steganography aims to hide information (hyperlinks, images, videos, etc.) within different information carriers such as image, video, and audio~\cite{yang2019hiding}. The classical steganography methods involve altering the least significant bits(LSBs)~\cite{wolfgang1996watermark,pevny2010using} and transforming domain techniques~\cite{chandramouli2003image_steg, wang2004cyber,qian2015deep,husien2015artificial,bi2007robust}.

In recent years, deep neural networks are utilized in image steganography algorithms~\cite{hayes2017generating,tang2017automatic}. The network is trained jointly to both encode and decode a message inside a cover file. Furthermore, the embedding image may be disturbed by a series of operations such as image cropping or compression between encoding and decoding steps.

Our methods require embedding 128-channel information into a JPEG image. Previous work does not have such a huge capacity or can not handle image corruption while embedding much data. HIDDEN~\cite{zhu2018hidden} proposes the first end-to-end trainable framework for data hiding. ~\cite{baluja2017hidingimage} tried to hide a color image within another same size image using deep neural networks. StegaStamp~\cite{2019stegastamp} attempts to encode and decode arbitrary hyperlink bitstrings into photos, and ~\cite{hu-2020-mononizing} tackles to embed one of the stereo images into another. Moreover, ~\cite{zhu2020video} embeds eight neighboring frames into one PNG image but can not deal with image disturbance. None of these methods has the capacity to embed 32-layer MPIs (128 channels) into a single 3-channel JPEG image without noticeable perceptual artifacts. 

The most significant difference between our method and the steganography methods is the primary purpose. In our method, the embedding image is designed to embed MPI while preserving a pleasant appearance without evident artifacts, but not to hide undetectable information.

\section{Method}
The overall framework of the proposed method is shown in Figure~\ref{fig:overview}. The input of the network is an MPI sequence consisting of 32 layers and one reference image. The MPI sequence is denoted as $M = \{m_i | i = 0, 1, ...31\}$, where $i$ denotes the MPI plane index. The shape of each MPI plane $m_i$ is $H \times W \times 4$. $H, W$ refers to the height and width of images. Each MPI plane $m_i$ is a 4 channel RGBA layer consisting of a color image $c_i$ and a alpha/transparency layer $\alpha_i$. And the reference image $I_{ref}$ is a $H \times W \times 3$ RGB color image.

Our framework consists of an embedding network to encode the MPI into a single JPEG image, a restoration network to reveal embedded information, and a discriminator to distinguish whether the image has weird artifacts. 

\begin{figure*}[t!]
\centering
\includegraphics[width=1\linewidth]{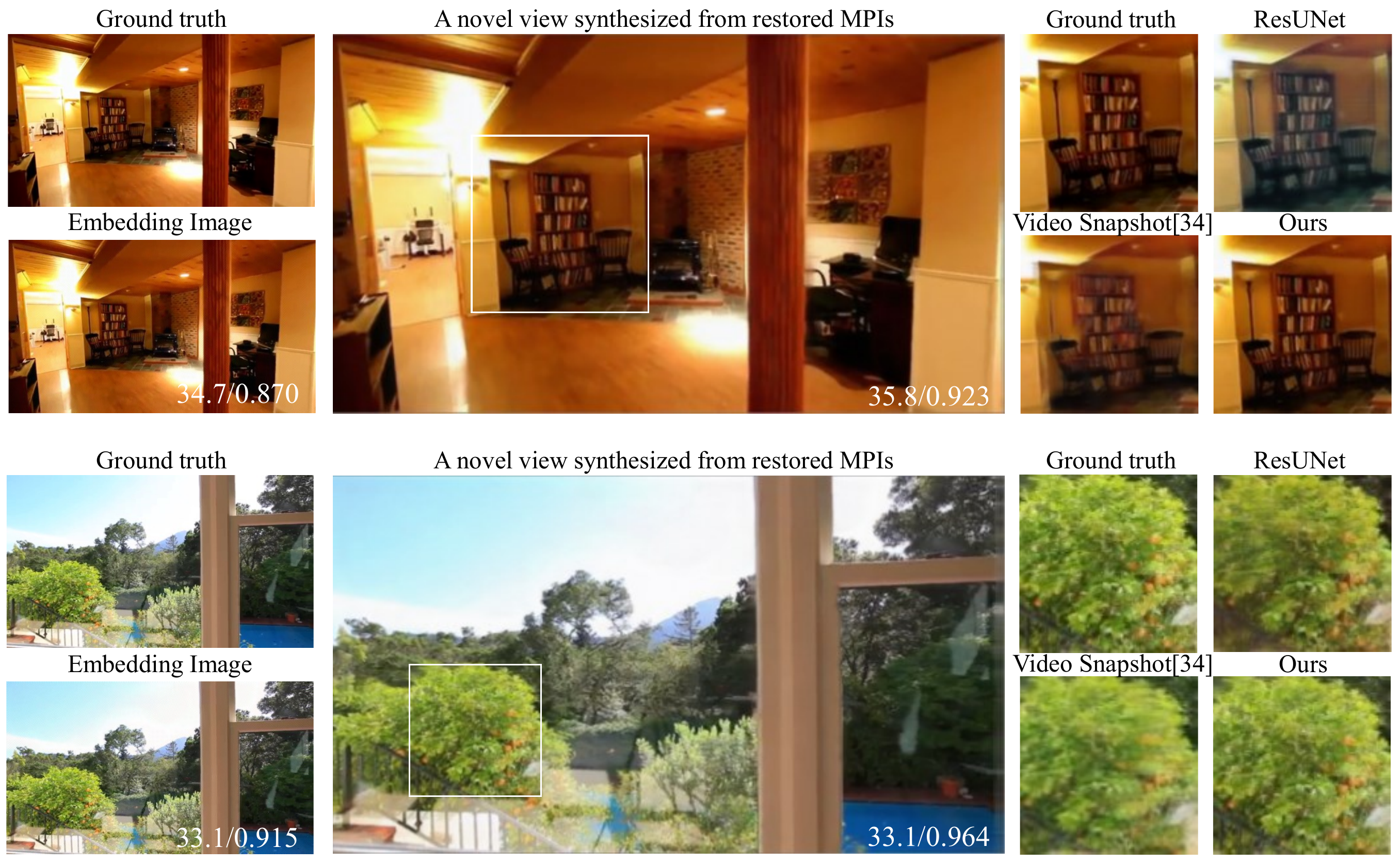}
\caption{The results of the embedding images and the synthesized images. The PSNR/SSIM scores are labeled on the images.
}
\label{fig:compare_render}
\end{figure*}

\subsection{Embedding Network}
The embedding network takes a sequence of 32 layers from an MPI $M$ and a reference image $I_{ref}$ as its input. The MPI is generated using state-of-the-art novel view synthesis methods~\cite{zhou2018stereo,PB}. We adopt two frameworks to demonstrate our generalization ability to handle different types of MPI representation.

Since the complexity and high channel amount of MPIs, it is difficult to directly use a basic encoder to embed the content in MPIs into images. We use three feature extractors $E_{rgb}$, $E_{\alpha}$, $E_{ref}$ to extract the features from RGB layers, alpha layers and the reference image respectively.

For $E_\alpha$ and $E_{ref}$, we use two convolution layers to extract features $S_{\alpha}$, $S_{ref}$. All the $\alpha_i$ are concatenated together then fed into $E_\alpha$. 
In $E_{rgb}$, for each input $c_i$, we extract features $s_i$ by an independent branch consisting of two convolutional layers. The branches do not share weights because the importance of each color layer is different when rendering new views. 

\paragraph{Feature Fusion.}
To fuse the 32 feature maps $s_i$ of each RGB layer $c_i$, we design a fusion mechanism with the alpha layers $\alpha_i$. Since the alpha layer represents the transparency of each RGB layer $c_i$, we fuse the features $\{s_0,...s_{31}\}$ based on alpha layers:
\begin{equation}
    S_{rgb} = \sum_{i=0}^{31} \alpha_i \times s_i.
\end{equation}

Then we concatenate $S_{rgb}$, $S_{\alpha}$, $S_{ref}$ and feed it into two convolutional blocks for downsampling, four residual blocks~\cite{he2016deep} and two upsampling layers implemented by bilinear interpolation and convolutional blocks. Skip connections across layers are used to preserve low-level details.

To make the model robust for real-world scenarios, we apply a series of differentiable image transformations on the embedding image to approximate the operations that social media users may perform. 
\paragraph{JPEG Compression.}
Photos are usually compressed in a lossy format when shared online. The embedding image is first quantized into an 8-bit image by the differentiable rounding operation. Then the JPEG compression is implemented by computing the discrete cosine transform of each $8 \times 8$ block in an image. The resulting coefficients are quantized by rounding to the nearest integer. This rounding step is not differentiable, so we follow the operation proposed by Shin \etal~\cite{jpeg} to approximate the rounding step:
\begin{align}
    &Q(x) = round(x), \\  
    &Q'(x) = 1,
\end{align}\label{quan}where $Q(x)$ is used in forward pass, meanwhile, the gradient $Q'(x)$ is set to be 1 in the backward propagation. The JPEG quality is empirically set as 90. Our JPEG compression is equivalent to the standard JPEG compression but differentiable. The forward pass of our JPEG compression is the same as the standard JPEG compression.

\paragraph{Image Manipulation.}
When users distribute a photo on social media, they may apply some photo filters for a better appearance before sharing the photo. Thus, we perform a series of random affine color transformations~\cite{2019stegastamp}, including brightness, contrast, hue, and saturation adjusting, to approximate this operation. Besides, to make our model robust to image cropping, we apply random cropping on the embedding images.

\begin{figure*}[t!]
\hspace{-4mm}
\setlength\tabcolsep{2pt}
\begin{tabular}{ccccc}
 \includegraphics[width=0.2\linewidth,height=3.1cm]{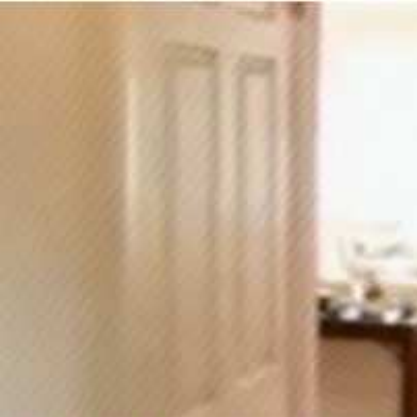}  & 
 \includegraphics[width=0.2\linewidth,height=3.1cm]{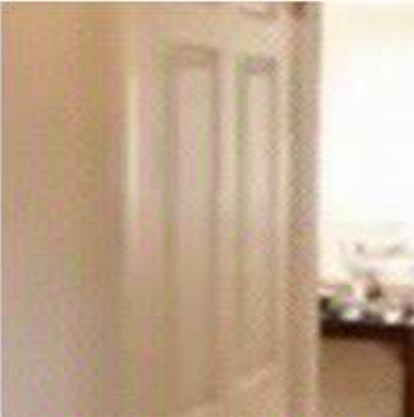}  & 
 \includegraphics[width=0.2\linewidth,height=3.1cm]{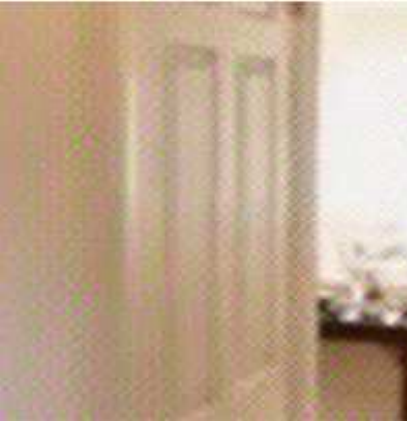} &
\includegraphics[width=0.2\linewidth,height=3.1cm]{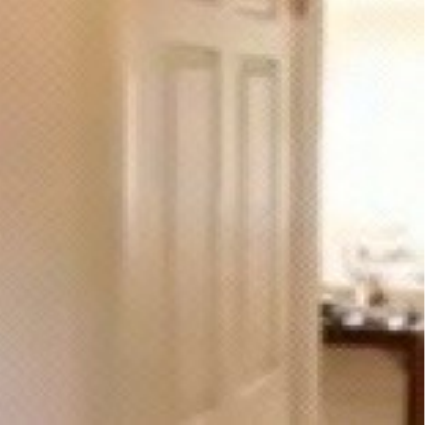} &
\includegraphics[width=0.2\linewidth,height=3.1cm]{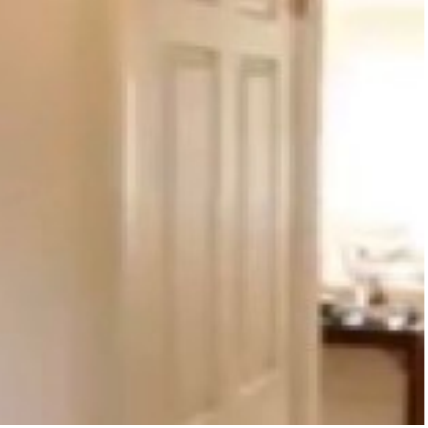}\\
Video Snapshot~\cite{zhu2020video} & Without frequency  & Without adversarial & Full model & Ground truth \\
\end{tabular}
\caption{The comparison shows the effect of loss functions. 
Without the frequency loss and the adversarial loss, the embedding image has evident color strips and high-frequency artifacts.
We recommend readers zoom in for the details.}\label{fig:loss}
\end{figure*}

\subsection{Restoration Network}
The input of the restoration network is the embedding image $\tilde{I}_{e}$.
Following~\cite{zhu2020video}, the architecture of the restoration network consists of eight residual blocks, one flat convolution layer, and skip connections across different layers. The output of the restoration network is the restored 32-layer MPIs.

Since it is intractable to embed all the details of the MPIs (128 channels) perfectly into JPEG images while preserving the visual perceptual quality, we force the network to embed the most important content for novel view synthesis.
Thus, we apply a random render module as guidance to the network. 

\subsection{Discriminator}
Because the information amount to be embedded is quite large, the embedding network will generate embedding images within highly-evident high-frequency artifacts for the purpose of restoring the embedded information. Thus, we adopt adversarial training to force the embedding image to be close to the reference image. The image discriminator is used to detect whether the embedding image has unnatural artifacts. The building block of the discriminator is multi-scale PatchGAN~\cite{pix2pix2017}.

\begin{figure*}[t!]
\centering
\includegraphics[width=1\linewidth]{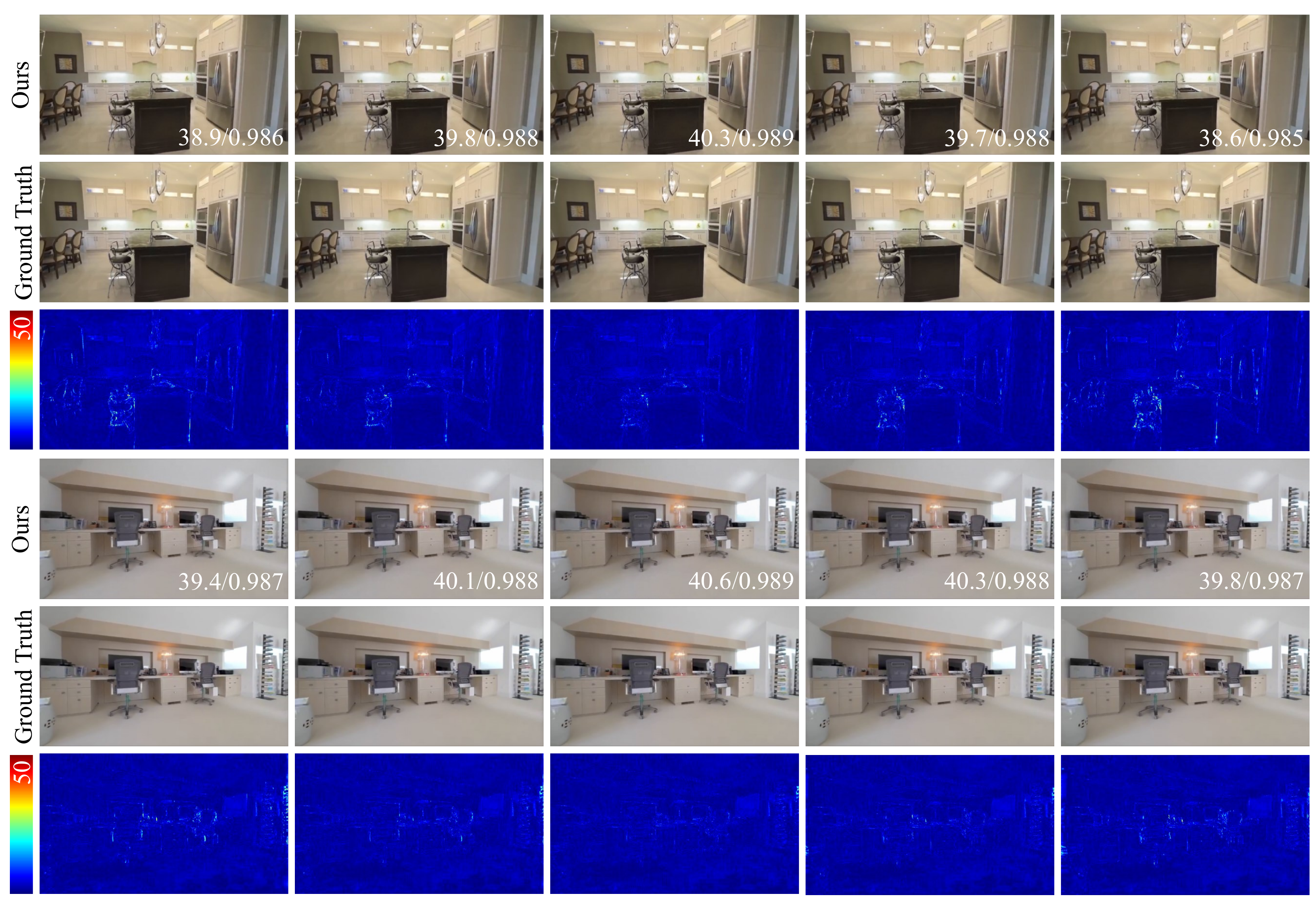}
\caption{The results of the synthesized images and the difference maps. The PSNR/SSIM scores are labeled on the images.
}
\label{fig:compare_render}
\end{figure*}

\subsection{Loss function}
The embedding network, restoration network, and discriminator are jointly trained. The overall loss function for the embedding network and the restoration network is a linear combination of multiple loss functions: 
\begin{equation}
\mathcal{L}_{G} = \lambda_1\mathcal{L}_{reg} + \lambda_2\mathcal{L}_{P} + \lambda_3\mathcal{L}_{freq} + \lambda_4\mathcal{L}_{res} + \lambda_5\mathcal{L}_{ren},
\end{equation}
where $\lambda_1, \lambda_2, \lambda_3, \lambda_4, \lambda_5$ are set as 8, 6, 0.003, 30, 1.

\noindent\textbf{Frequency domain loss.} We introduce a novel frequency-domain loss to suppress the network to embed information using evident high-frequency patterns. To our best knowledge, we are the first to use a frequency domain loss to help suppress evident artifacts. This is motivated by our observation that, when embedding a large amount of information, like several images~\cite{zhu2020video} or MPIs, the network tends to use high-frequency color strips to embed information. These artifacts are evident and visually unpleasing.
We convert the embedding image and reference image to the frequency domain by Fast Fourier Transforms (FFT) and enforce them close in the frequency domain:
\begin{equation}
    \mathcal{L}_{freq} =  ||FFT(\tilde{I_e}) - FFT(I_{ref}) ||^2,
\end{equation}
where $\tilde{I}_e$ is the embedding image, and $I_{ref}$ is the reference image. With this loss function, the color strip artifacts are effectively suppressed.

\noindent\textbf{Adversarial loss.} 
We use a discriminator to enforce the embedding image perceptually close to a natural image:
\begin{equation}
    \mathcal{L}_{D} = log(1 - \mathcal{D}(\tilde{I}_e)) + log\mathcal{D}(I_{ref}).
\end{equation}

\noindent\textbf{Regularization loss.}
$\mathcal{L}_{reg}$ penalizes the discrepancy between the reference image and embedding image.
\begin{equation}
    \mathcal{L}_{reg} = || \tilde{I}_e - I_{ref} ||^2,
\end{equation}

\noindent\textbf{Perceptual loss.}
We also use the perceptual loss~\cite{ChenKoltun2017} between $I_{ref}$ and $\tilde{I}_{e}$. We use VGG19 model~\cite{vgg} for feature extraction and define a $L_1$ loss between embedding image and reference image in the feature domain:
\begin{equation}
\mathcal{L}_{P} =  \sum_{j=1}^{n}\frac{1}{N_j}\left \| \Phi_j(\tilde{I}_e) -\Phi_j(I_{ref}) \right \|_1 ,
\end{equation}
where $n$ is the number of VGG feature layers. $\Phi_j$ denotes the feature map from the $j$-th layer in the VGG-19 network, and the number of parameters of $\Phi_j$ is $N_j$.

\noindent\textbf{Render loss.} Since the data amount need to be embedded is quite large, we use a render loss to encourage the system to represent the most important features. The render loss consists of a MSE loss and a perceptual loss :
\begin{align}
&\mathcal{L}_{ren} = \lambda_{rmse}\mathcal{L}_{rmse} + \lambda_{rp}\mathcal{L}_{rp}, \\
&\mathcal{L}_{rmse} = || R_\theta(\tilde{M}) - R_\theta(M)||^2,  \\
&\mathcal{L}_{rp} = \sum_{j=1}^{n}\frac{1}{N_j}\left \| \Phi_j(R_\theta(\tilde{M})) -\Phi_j(R_\theta(M)) \right \|_1, 
\end{align}
where $\theta$ represents random render parameters. $R$ refers to the render process~\cite{zhou2018stereo} of reprojecting the MPI to the coordinate system of the target view, then compositing the reprojected RGB layers from back to front according to the corresponding alphas. And $\tilde{M}$ represents the restored MPIs. The translation parameters is sampled uniformly from $U[-0.5,0.5]$. The rotation parameters is sampled from $U[-8^{\circ},8^{\circ}]$ degrees. $\lambda_{rmse}$ and  $\lambda_{rp}$ are set as 100 and 15.

\noindent\textbf{Restoration loss.} We use a restoration loss to measure the similarity between restored MPIs and ground-truth MPIs:
\begin{equation}
    \mathcal{L}_{res} = \sum_i  \lambda_{rgb} || \tilde{c}_i \odot \alpha_i - c_i  \odot \alpha_i||^2 + ||\tilde{\alpha}_i - \alpha_i ||^2, 
\end{equation}
where $i$ is the index of the planes of MPIs, $c_i$, $\alpha_i$ are the ground-truth color image and alpha image at index $i$. $\tilde{c}_i$, $\tilde{\alpha}_i$ are the corresponding predicted color and alpha image. $\lambda_{rgb}$ is set as 10. Thus, the network is enforced to emphasize the pixels within higher alpha values.

The overall training loss is $\underset{\mathcal{G}}{\mathrm{min}}\big(\underset{\mathcal{D}}{\mathrm{max}}\mathcal{L}_D + \mathcal{L}_{G}\big)$.

\begin{table*}
\centering
\resizebox{1\linewidth}{!}{
\ra{1.25}
\begin{tabular}{@{}lccccccccccccc@{}}
\toprule
& \multicolumn{6}{c}{Stereo-Mag~\cite{zhou2018stereo}} && \multicolumn{6}{c}{PB-MPI~\cite{PB}} \\
\cmidrule(l{3mm}r{3mm}){2-7} \cmidrule(l{3mm}r{3mm}){9-14}
& \multicolumn{3}{c}{Embedding} & \multicolumn{3}{c}{Render} && \multicolumn{3}{c}{Embedding} & \multicolumn{3}{c}{Render} \\
& SSIM$\uparrow$  & PSNR$\uparrow$ & LPIPS$\downarrow$  & SSIM$\uparrow$  & PSNR$\uparrow$ & LPIPS$\downarrow$ && SSIM$\uparrow$  & PSNR$\uparrow$ & LPIPS$\downarrow$ & SSIM$\uparrow$  & PSNR$\uparrow$ & LPIPS$\downarrow$ \\ 
\midrule
UNet & 0.8830 & 25.644 & 0.2654 & 0.8490 & 22.676 & 0.2717 && 0.8619 & 26.005 & 0.2783 & 0.8016 & 21.869 & 0.3057 \\
ResUNet & 0.8229 & 25.780 & 0.3763 & 0.8931 & 24.326 & 0.2169 && 0.7871 & 25.565 & 0.3785 & 0.8441 & 23.071 & 0.2614 \\
Video Snapshot~\cite{zhu2020video} & 0.7939 & 30.395 & \textcolor{blue}{\textbf{0.1007}} & 0.8703 & 27.066 & 0.1732 && 0.7705 & 30.305 & 0.1045 & 0.8417 & 25.885 & 0.2553 \\
\midrule
Ours w/o GAN  & 0.8661 & 32.884 & 0.1832 & 0.9664 & 35.174 & 0.0637 && 0.9537 & 36.529 & 0.1215 & 0.9393 & 30.435 & 0.1107 \\
Ours w/o Render  & \textcolor{red}{\textbf{0.9688}} & \textcolor{blue}{\textbf{33.998}} & \textcolor{red}{\textbf{{0.0926}}} & 0.9492 & 32.301 & 0.1137 && \textcolor{red}{\textbf{0.9695}} & \textcolor{red}{\textbf{37.650}} & \textcolor{red}{\textbf{0.0667}} & 0.8912 & 27.497 & 0.1845 \\
Ours w/o Frequency  & 0.7970 & 30.911 & 0.2844 & \textcolor{red}{\textbf{0.9773}} & \textcolor{red}{\textbf{38.105}} & \textcolor{red}{\textbf{0.0509}} && 0.7983 & 27.796 & 0.2349 & \textcolor{blue}{\textbf{0.9513}} & \textcolor{blue}{\textbf{30.831}} & \textcolor{red}{\textbf{0.0948}} \\
\hline
Ours         & \textcolor{blue}{\textbf{0.8941}} & \textcolor{red}{\textbf{34.616}} & 0.1736 & \textcolor{blue}{\textbf{0.9750}} & \textcolor{blue}{\textbf{36.683}} & \textcolor{blue}{\textbf{0.0535}} && \textcolor{blue}{\textbf{0.9593}} & \textcolor{blue}{\textbf{36.736}} & \textcolor{blue}{\textbf{0.0951}} & \textcolor{red}{\textbf{0.9533}} & \textcolor{red}{\textbf{32.840}} & \textcolor{blue}{\textbf{0.0951}} \\
\bottomrule
\end{tabular}
}
\vspace{2mm}
\caption{The comparison of the quality of the embedding images and the rendered novel views. Our method surpasses the baseline methods by a large margin. The ablation study shows the effectiveness of different loss functions. Our full model achieves the best balance between the quality of embedding images and the rendered images. The best and second-best scores are indicated in \textcolor{red}{\textbf{red}} and \textcolor{blue}{\textbf{blue}}.}
\label{table:hide_quality}
\end{table*}

\section{Experiments}
We choose two novel view synthesis methods, Stereo-Mag~\cite{zhou2018stereo} and PB-MPI~\cite{PB}. Our network is supervised by ground-truth MPIs generated by these two methods. There are 128 MPI planes in the output of the original PB-MPI. To make it fit our method, we first convert the output of PB-MPI into 32 MPI planes by merging four adjacent MPI planes as one. We compare our approach with other approaches on the quality of the embedding image and the quality of restored information. Extensive experiments and details are presented in the supplementary material.

\subsection{Datasets}
We conduct our experiment on the RealEstate10K dataset~\cite{zhou2018stereo}. The RealEstate10K dataset contains about 10,000 YouTube videos of indoor and outdoor real estate scenes. We generate training samples by random sampling frames during training and conduct data augmentation by randomly crop patches. During testing, we randomly select 1500 sequences and use frame $10^{th}$ and frame $14^{th}$ as source frames. We conduct experiments with images of resolution  $512 \times 288$. To compare the quality of the rendered images, we render novel views for each scene at nine poses and compute the average metric score as render score.


\subsection{Evaluation metrics}
We evaluate our model using several metrics measuring the quality of embedding images and rendered novel views. We use SSIM, PSNR, and learned perceptual image patch similarity (LPIPS)~\cite{lpips}. Higher SSIM, higher PSNR, and lower LPIPS distances suggest better performance. 

\subsection{Baselines}
From the first perspective, to evaluate the performance of embedding information, we compare our model with several baselines. Some are state-of-the-art approaches, and the rest are variants of our model. From the second perspective, to demonstrate that our performance is better than simply applying the novel view synthesis model on a single image, we compare our method with several single image view synthesis methods.

\paragraph{UNet and  ResUNet with MSE loss.} 
Since the UNet~\cite{ronneberger2015unet}, ResUNet, and their variants are commonly used in related tasks like~\cite{2019stegastamp, zhu2020video}, we implement these two network architectures as naive solutions for this task. The embedding network and restoration network are implemented using UNet and ResUNet. Both of these two methods are trained using MSE Loss. We demonstrate that the network structure to embed such a large amount of information should be carefully designed.

\paragraph{Video Snapshot~\cite{zhu2020video}.}
Video Snapshot is the most relevant state-of-the-art work that embeds eight consecutive frames into one image. We treat each MPI layer as a frame and cautiously reimplement this work.

\paragraph{WG, WR, and WF.} We evaluate the performance of our model without GAN training (WG), render module (WR), frequency domain loss (WF) separately to validate the effectiveness of each module.

\paragraph{SynSin~\cite{wiles2020synsin} and Single-View View Synthesis (S-MPI)~\cite{single_view_mpi}.} SynSin and S-MPI are state-of-the-art single image view synthesis methods. We compare both the quality and running speed between our method and them to show the advantage of our method over the single image view synthesis method.

\begin{table}[t!]
\centering
\setlength{\tabcolsep}{3mm}
\ra{1.05}
\begin{tabular}{lccc}
\toprule
& \multicolumn{3}{c}{Render} \\
\midrule
& SSIM$\uparrow$ & PSNR$\uparrow$ & LPIPS$\downarrow$ \\
\midrule
SynSin~\cite{wiles2020synsin}      & 0.7851 & 24.078 & 0.2016   \\
S-MPI~\cite{single_view_mpi}   & 0.8084 & 23.837 & 0.1905  \\
\midrule
Ours & {\textbf{0.8810}} & {\textbf{26.732}} & {\textbf{0.1507}} \\
\bottomrule
\end{tabular}
\vspace{1mm}
\caption{We compare our method with state-of-the-art single image view synthesis methods. The results demonstrate that embedding the novel views into a single image produces much better results than single image view synthesis.}
\label{table:render_quality}
\end{table}

\subsection{Evaluation}
As shown in Fig.~\ref{fig:compare_render}, the novel views rendered from the restored MPIs have high visual quality. The difference between rendered novel views and ground truth is nearly imperceptible.
A comprehensive comparison between our methods and other methods is shown in Table~\ref{table:hide_quality}. For both the MPIs predicted by the Stereo-Mag and MPI-PB, the performance of our model surpasses that of naive baselines and Video Snapshot~\cite{zhu2020video} on the image quality of the embedding images and rendered novel views by a large margin. 
In Stereo-Mag, in terms of SSIM and PSNR, the performance of our embedding image (0.8941, 34.616) is far higher than that of Video Snapshot (0.7939, 30.395). Moreover, the quality of our rendered novel views (0.9750, 36.683) can significantly surpass that of Video Snapshot (0.8703, 27.066).
These statistical results quantitatively demonstrate the effectiveness of our network architecture and proper loss functions. 

For the ablation study, there is a trade-off between the perceptual quality of the embedding image and the accuracy of rendered views. When training without the discriminator and frequency domain loss, the embedding images degrade and appear to have apparent artifacts, as shown in Fig.~\ref{fig:loss}. It demonstrates that these two loss functions suppress the artifacts in the embedding images effectively. As shown in Table~\ref{table:hide_quality}, when training without the render module, the performance of rendering degrades. It is because the render module provides an accurate emphasis on regions critical for rendering. 
In most cases, our full model achieves the top scores, obtaining the best compromise between embedding quality and rendering perceptual performance.

Towards the same purpose of presenting social users with 3D photographs, we compare our method with the LDI~\cite{Shih3DP20} and the ``Two views''. The ``Two views'' embeds another view into the reference image using Mono3D~\cite{hu-2020-mononizing} and then renders novel views from the restored images. This evaluation is conducted in PNG format. 

\begin{table}[H]
\centering
\setlength{\tabcolsep}{1mm}
\begin{tabular}{lccc}
\toprule
& SSIM/PSNR & Speed (s) & Model (M) \\
\midrule
LDI   & 0.8426/25.735  & 42.3 & 438  \\
\midrule
Two views (PNG) & 0.9016/27.639 & 0.448 & 264\\
\midrule
Ours (PNG) & 0.8953/27.198 & 0.017 & 6.3 \\
\bottomrule
\end{tabular}
\vspace{1mm}
\caption{The comparison between our method and LDI~\cite{Shih3DP20} in terms of render quality.}
\label{table:compare}
\end{table}

Our model renders fast and uses a small decoder network while maintaining high rendering quality. When compared with LDI, the advantage of embedding MPI is that MPI can store content in occlusion areas. In contrast, occlusion areas are inpainted by the neural network in the LDI and thus often less realistic. Our model is generally stable, while the performance of the LDI depends on the image content. The rendering of the ``Two views'' is slow, and its model size is large since the ``Two views'' method needs to compute the MPI first for rendering. 

\begin{figure}[t!]
\centering
\includegraphics[width=0.49\linewidth]{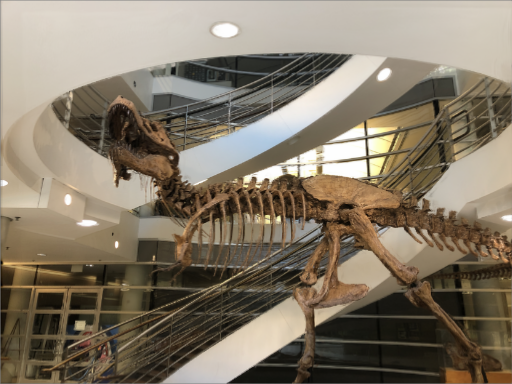} 
\includegraphics[width=0.49\linewidth]{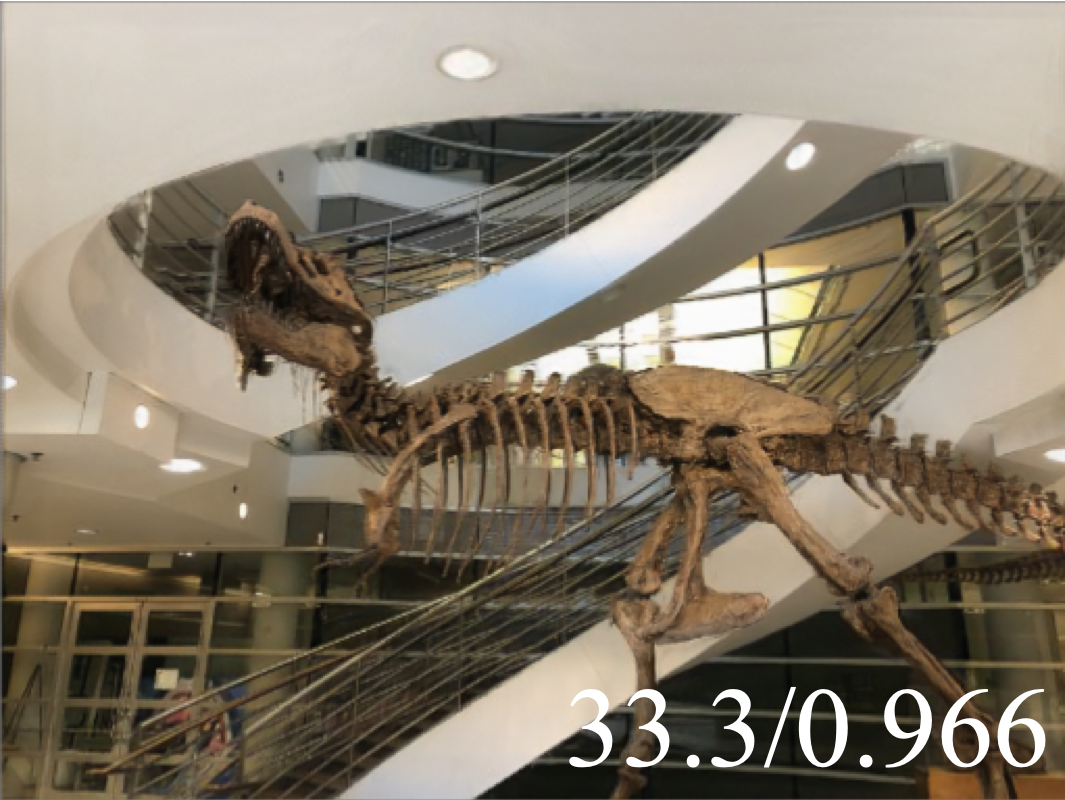}
\vspace{1mm}
\caption{The visual results on the LLFF dataset. The left is the ground truth, and the right is a predicted novel view.
}
\label{fig:llff}
\end{figure}

We conduct an additional experiment on MPI generated by LLFF~\cite{mildenhall2019llff}. Since the dataset provided by  LLFF is too small to train on directly, we slightly fine-tune our Stereo-Mag~\cite{zhou2018stereo} model on a small proportion of the LLFF dataset. As shown in Figure~\ref{fig:llff}, our model can achieve visually pleasing results on the LLFF dataset. Therefore, our method can be adaptive to most MPI-based methods.

We also compare our method with the single image view synthesis methods, SynSin~\cite{wiles2020synsin}, and S-MPI~\cite{single_view_mpi}. As shown in Table~\ref{table:render_quality}, our method achieves better image rendering performance. Furthermore, the inference time of our method is much less than that of the single image view synthesis methods. 

Table~\ref{table:size_and_speed} depicts a comparison between different methods in terms of model sizes and inference time. Our method only takes 0.003 seconds to infer an MPI from an embedding image, which makes it possible to render novel views in real-time on a regular smartphone. 

\begin{table}[t!]
\centering
\setlength{\tabcolsep}{1mm}
\ra{1.05}
\begin{tabular}{@{}cccc@{}}
\toprule
& Model (M) & Recover MPI (s) & Render (s)\\
\midrule
SynSin~\cite{wiles2020synsin}  & 273 & -    & 0.077 \\
S-MPI~\cite{single_view_mpi} & 167 & 0.692  & 0.706 \\
Stereo~\cite{zhou2018stereo} & 185 & 0.427  & 0.441 \\
PB~\cite{PB}                 & 524 & 2.483  & 2.497 \\
\midrule
Ours & {\textbf{6.3}} & {\textbf{0.003}} & \textbf{0.017}\\
\bottomrule
\end{tabular}
\vspace{1mm}
\caption{The first column is the model size. The second column is the inference time for generating single MPI. The third column is the time for rendering (including MPI generation) a single view.}
\label{table:size_and_speed}
\end{table}

The experiments of the robustness of our model against image manipulations are presented in supplementary material.

\section{Conclusion}
We propose a novel approach for embedding novel views in a single JPEG image. Comprehensive experiments are conducted on different datasets and different MPI prediction methods. The result shows that our method can recover high-fidelity novel views from a slightly modified JPEG image. Furthermore, the experiments show that the proposed method is robust to image editings like cropping and color adjusting. Since the decoder in our framework is a lightweight convolution network, it can be deployed on regular smartphones.

\clearpage
\newpage
\balance

{\small
\bibliographystyle{ieee_fullname}
\bibliography{egpaper_final}
}

\end{document}